# An approach for mistranslation removal from popular dataset for Indic MT Task

Sudhansu Bala Das[1], Leo Raphael Rodrigues[1], Tapas Kumar Mishra[1], Bidyut Kr. Patra[2]

[1*] National Institute of Technology(NIT), Rourkela, Odisha, India.
[2] Indian Institute of Technology (IIT), Varanasi, Uttar Pradesh, India.

Corresponding author(s). E-mail(s): baladas.sudhansu@gmail.com;

**Abstract**

The conversion of content from one language to another utilizing a computer system is known as Machine Translation (MT). Various techniques have come up to ensure effective translations that retain the contextual and lexical interpretation of the source language. End-to-end Neural Machine Translation (NMT) is a popular technique and it is now widely used in real-world MT systems. Massive amounts of parallel datasets (sentences in one language alongside translations in another) are required for MT systems. These datasets are crucial for an MT system to learn linguistic structures and patterns of both languages during the training phase. One such dataset is Samanantar, the largest publicly accessible parallel dataset for Indian languages (ILs). Since the corpus has been gathered from various sources, it contains many incorrect translations. Hence, the MT systems built using this dataset cannot perform to their usual potential. In this paper, we propose an algorithm to remove mistranslations from the training corpus and evaluate its performance and efficiency. Two Indic languages (ILs), namely, Hindi (HIN) and Odia (ODI) are chosen for the experiment. A baseline NMT system is built for these two ILs, and the effect of different dataset sizes is also investigated. The quality of the translations in the experiment is evaluated using standard metrics such as BLEU, METEOR, and RIBES. From the results, it is observed that removing the incorrect translation from the dataset makes the translation quality better. It is also noticed that, despite the fact that the ILs-English and English-ILs systems are trained using the same corpus, ILs-English works more effectively across all the evaluation metrics.

## 1 Introduction

Machine Translation (MT) is a modern technology that simplifies the translation of text or speech among languages, allowing for smooth interaction as well as breaking down language barriers. With increased global interdependence and interactions between cultures, MT is becoming more essential for promoting communication as well as fostering collaboration among individuals who speak various languages. Nonetheless, accomplishing precise and intact translations is still a crucial challenge when the dataset comes into play. A "dataset" is an extensive database of bilingual or multilingual texts utilized as training data for MT systems. It consists of pairs of interpretations from the source language to the target language. To examine these datasets, different MT models are used, which learn the textual structures and patterns in order to do the translation more accurately. However, the quality of the dataset used to train a Machine Translation (MT) system has a significant effect on the precision as well as the fluency of the output. When the dataset is of poor quality, with inaccuracies, inconsistencies, or irrelevant data, the MT system's performance will suffer, resulting in poor translations or even nonsensical outputs. Hence, the massive qualitative dataset is important for the MT System.

In multilingual countries such as India, the availability of massive, high-quality datasets for Indian languages (ILs) is important for establishing an effective MT system. But it is observed that several ILs have constrained or inadequate datasets, which makes it challenging to establish robust MT models for those languages. However, researchers have come up with a massive parallel dataset collection for ILs, i.e., the Samanantar Dataset [1]. But as this dataset has been collected from a variety of sources, it consists of an immense quantity of noise. The existence of noise in the dataset has an adverse effect on the efficiency and precision of the MT task. As a result, there are no MT models on ILs that can generate flawless translations (precisely convey the meaning and actual translation from source to target language).

As MT becomes a more prominent mode of translation, the quality of its translations becomes increasingly important, such that it fulfills predetermined standards and captures the tone and content of the source text as accurately as possible. Hence, a high-quality dataset becomes an essential resource for addressing these issues, with a significant effect on the efficiency of the MT system. A novel approach to MT called Neural Machine Translation (NMT) seeks to maximize translation performance by offering an originally trainable neural network model [2], [3]. Compared to conventional MT systems such as Rule-Based [6], Example-Based [8], and Statistical Based [9], [13], NMTs are much simpler to construct and train. In addition, these systems are able to deliver cutting-edge results for translation tasks in various language pairs [11]. The size of a dataset utilized for training an NMT system has a substantial influence on its performance [23]. The more comprehensive the dataset, the greater is the quality of the output generated by the model [33].

Many researchers working with ILs have recently built various MT systems and models to experiment with various approaches to translate ILs to ENG and vice versa, but they are still unable to achieve sufficient accuracy. The translation quality of models built on ILs can vary according to their dataset quality and quantity. Therefore, the individual relevance of the quality and quantity of the dataset during training remains an open question. In order to encourage the growth of MT for ILs, performance evaluation of the translation quality of machine-generated output becomes increasingly important. In this paper, the Samanatar dataset is used for the analysis of the role of different filtration techniques in the MT system. The following is the summary of the significant contributions of this paper

- In this paper, Hindi (HIN) and Odia (ORI) language dataset is considered for the experiment. All the translations are evaluated using standard evaluation metrics i.e. BLEU, RIBES, and METEOR. • A baseline system using NMT for 2 ILs is developed and examined in this paper.
- The paper also proposes a novel technique to remove incorrect translations from the dataset and checks its effect on MT tasks.
- It also investigates how different dataset divisions (into multiple subsets) influence the translation quality.
  This paper is arranged as follows: Section 2 focuses on some prominent work on machine translation. The dataset and languages used for our experiment are illustrated in Section 3. Section 4 discusses our methodology, which includes the algorithm proposed for the removal of mistranslations from the dataset. Results are illustrated in Section 5, and Section 6 gives the conclusion and future work.

## 2 Related Work

When attempting to obtain parallel training datasets for low or high-resource languages, gathering datasets, such as those found in Samanantar [1] is a good place to start work on any MT tasks. Samanantar dataset contains a large number of different Indian languages (ILs) with varying sizes, collected from various sources (e.g., Wikipedia articles, software handbooks, and religious texts). After collecting the dataset, the next step is to increase the quality of the dataset. But while examining it, it is also noticed that it contains too much noise. Many researchers have examined the issues of data cleaning and selection for MT works [20], [21]. Some research works [25] have discussed assessing and improving the self-training properties in an NMT (via selective, high-uncertainty sampling) using solely monolingual data, while others [26] illustrated assessing the monolingual data quality for NMT tasks. Understanding the effect of the quality of parallel datasets is yet to be explored.

One of the most effective and simple approaches for enhancing the capabilities of MT systems using monolingual data is the self-learning approach [17]. This technique involves translating a substantial amount of data from the source language into a new or existing model and then using the decoded parallel data to enhance the quality of the base model. But as per Guillem et al [34], having too much training data does not necessarily ensure a better model. Their method has been shown to skip training on the generated output of the additional data or use quality estimation [18], or data extraction systems to pick only the most qualitative sentences in order to achieve the best performance. Jiao et al ([25]) and Wu et al ([26]) research conducts a largescale human assessment of a standard dataset and discovers serious quality concerns, particularly in low-resource languages.

The impact of uncertainty in the NMT model-fitting has been examined in the experiment shown by Ott et al [31]. Although the model is usually well-calibrated at both the token and sentence levels, it tends to spread probability mass too widely. After examining some of the consequences of this mismatch, it is observed that the application of excessive probability results in poor-quality model samples.

In the year 2018, Kharallah et al [7] investigated how different types of noise in parallel training data affect the overall performance of neural machine translation systems. They generate five types of artificial noise and explore how they affect neural and statistical machine translation performance. According to the findings, noise harms neural models more than statistical models.

For enhanced NMT performance, monolingual data can be used and examined [26]. The method starts with creating synthetic bitext by translating monolingual data from one domain to the other using models trained on real bitext. Next, the model is trained using a noisy version of the synthetic bitext which has random corruption in the source sequences. Finally, the model is fine-tuned using a clean version of the synthesized bitext without noise.

The paper by Muischnek et al [41] focuses on recognizing and addressing problems in data that affect NMT systems. The authors observe that dataset filtering, which is a pre-processing step to clean and improve the quality of the data, has a positive impact on NMT systems for Latvian and Estonian languages but not for Finnish. This suggests that the effectiveness of dataset filtering may vary depending on the language and the specific NMT systems in use.

According to Koehn et al [27], resilience to noise in MT can be modeled as a domain adaptation issue. The importance of their work has grown in the current years with more and more researchers using the approach to deal with noise, as seen in the establishment of a competition on dataset filtration and orientation in WMT tasks [28]. In the year 2022, different filtering techniques are examined by Frid et al [5]. Their work compares the performance of different data filtration tools in filtering out different types of noisy data. Their work compares the performance of different data filtration tools in filtering out different types of noisy data. They also look into the impact of filtration with these tools and how well the models perform in the downstream task of NMT by generating a dataset with a mix of clean and noisy data; then training NMT engines with the leading to the filtered dataset. In the next section, an overview of the dataset is explained.

## 3 Dataset and Languages Used

A dataset is a massive collection of text (often billions of words) developed by authentic language users and is used to investigate the role of phrases, words, and language in any MT task [14]. Dataset in MT can be classified into three types:

- Monolingual dataset: The most common type of dataset in MT is a monolingual dataset. It has texts in only one language. The IndicCorp dataset is an example of a monolingual dataset [16].
- Parallel dataset: A parallel dataset is made up of translation pairs in two languages. Both language translation pairs should be aligned, which means that corresponding segments, normally paragraphs or sentences, must be matched. The Samanantar dataset is a parallel dataset [1].
- Multilingual dataset: A multilingual dataset is a translation group of more than two languages, where each group is aligned. The Flores-200 dataset is a great example
  [15].

For our experiment, Samanantar dataset is used to compare and test the impact of dataset quality and quantity on MT tasks using Hindi (HIN) and Odia (ODI) languages. This dataset contains over 40 million sentence pairs from English (ENG) to ILs. The Flores200 dataset [15] is utilized for testing purposes.

- Hindi(HIN): As per Samanantar dataset, HIN with a dataset of 8.56M. Being a prominent official language of India, HIN is the mother tongue of more than 615 million people and also serves as a second language for more than 341 million people. It uses the Devanagari script and belongs to the Indo-European family.

- Odia(ORI): The official language of the Indian state of Odisha is Odia. From the dataset, it has been observed that ORI is treated as low resource language with a dataset of 1.00M. Subject-object-verb (SOV) is the format that is typically used for ORI. The Odia script follows from left to right

A rich morphological and structural variation can be seen in ILs, which makes MT and NLP tasks challenging [29]. In other words, these languages provide a wide range of interpretations and usages in terms of meaning and grammatical structure respectively. A significant structural difference between English and ILs is the word order. The subject-verb-object (SVO) sequence is used in English (ENG), whereas the majority of ILs primarily use the subject-object-verb (SOV) structure. In addition, English makes extensive use of the neuter gender. ILs also possess many features that English does not possess. For instance, many of them have verb genders and possess 8 grammatical cases, compared to ENG which has no gendered verbs and only 3 grammatical cases. Poor performance of IL translation with respect to ENG relative to other international languages can also be attributed to the loss of data that occurs during translation. ILs being able to contain more information than ENG results in a loss of context and information present in the ILs with the process of translation into English for its limit to represent all the informatic inputs. This effect is fairly pronounced in the case of translations from English to ILs, where the model does not have sufficient information to infer the additional context from English itself. Hence, in this work, we mainly focus to use our approach for the translation of English to 2 Indian languages and vice versa

## 4 Methodology

This section describes the algorithm proposed for mistranslation removal, the model overview, and the results obtained. The quality of data or a dataset cannot be guaranteed when obtained from different sources, which is crucial for an MT system to function effectively. For our experiments, we use both original unfiltered as well as after the removal of the mistranslation dataset. Dataset is standardized and tokenized with the Indic NLP library [36] for both Indian Languages. Previous research has shown that filtering techniques can have both positive and negative effects on training datasets [32]. Therefore, a novel technique is proposed in this paper using which we can remove the sentences with incorrect translations from the dataset.

**4.1 Removing Incorrect Translation: To check the quality of dataset**

While examining the dataset it is observed that it possesses incorrect translations. The term ''incorrect/mistranslations or incomplete translation'' in the context of a dataset refers to a translation that fails to accurately represent the intended significance of the original text. This can happen for various reasons, including translation errors, absence of contextual knowledge, or variations in the grammar and structure of the source and target languages. MT models acquire knowledge by analysing examples from the training dataset, and if the dataset has wrong or inaccurate translations, it can cause improper learning. As a result, the model is found to underperform in MT Tasks. These mistranslations can impede model performance.

It is critical to remove incorrect translations from a dataset in order to improve the accuracy and efficacy of MT models. To test the efficiency, the dataset is divided into multiple subsets in order to generate different corpus sizes. Then, on each subset, an extensive quality assessment is performed to identify and remove incorrect translations. A filtering technique to remove the mistranslations from the dataset without human intervention is proposed in this work. Although numerous metrics can be utilized to assess the quality of MT tasks, the BLEU (Bilingual Evaluation Understudy score) is widely used as it has been demonstrated to correlate well with human translation quality judgments.

BLEU score examines a machine-generated translation to one or more reference translations and assigns a score depending on how closely the machine-generated translation corresponds to the reference translations. This renders it a valuable metric to measure machine-generated translations comparable to reference translations and has been successfully demonstrated to be efficient at distinguishing between good and bad translations. As a result, for our experiments, the BLEU score is used to determine a threshold for the model in order to remove the incorrect translation. This threshold needs to be set according to the MT task's context and objective. The setting of the threshold defines the trade-off between retaining high-quality translations and maintaining a sizeable dataset. For our case, the threshold score is calculated using the validation dataset to remove incorrect translations from the training dataset. The threshold allows only the top 75% performant sentences to be retained in the dataset. The sentences with poor translation quality (BLEU score lesser than the threshold) are treated as noise and discarded from the dataset. The proposed method involves training a model on the original unfiltered dataset and the generation of the best checkpoint from the model. The best checkpoint is then used to create a translation of the validation dataset (used during training) which is then compared with its parallel counterpart to give us a BLEU score.

For example, when training an ENG-ORI model, the validation set's ENG portion is translated into ORI and then compared to the pre-existing ORI section of the validation set. This BLEU score is then used to create a threshold score (sentences below this score are perceived as mistranslations). The training dataset is then iterated over, and each ENG line is translated to its Indic counterpart (using the checkpoint) and compared to the original Indic line to extract a sentence BLEU score (using English to HIN/ORI model) and vice versa. If this score is below the calculated threshold scores, then those sentences having below that BLEU score are discarded from the dataset. For our experiment, the dataset is divided into three different sizes. Then, the effect for both these languages is investigated using two methods, namely Unfiltered Dataset and Removal of Mistranslated Sentences from the Dataset.

Case 1: Original Dataset.
Case 2: Quarter Dataset
Case 3: 500k Sentences from the Dataset (The size is chosen based on the relative size of the dataset available for both languages).

Following are the steps conducted to remove mistranslations from the dataset:

1. Train the model on the entire original unfiltered dataset (without removing any sentences).
2. Use the newly created model to translate the validation dataset and find its BLEU score using the existing validation dataset as a reference.
3. A threshold score (in this case, score/4) is created using this BLEU score.
4. The entire train set is translated using the existing model checkpoints.
5. The translated training set is compared with the original training set, and any sentences that receive a BLEU score lower than the computed threshold are eliminated from the original dataset.

Algorithm 1 describes the steps followed for our experiment. Tables 1 and 2 show the statistics of the dataset after filtering, as well as the removal of mistranslated sentences that were found in the dataset and the mistranslations that were noted in the dataset.

## 4.2 Model Training

The MT model is based on the transformer [35] architecture, with 6 encoder-decoder pairs and 512 hidden layers and token embeddings.

The models have been constructed **Algorithm** by the Fairseq [30] library. The dataset is segmented using the Byte Pair Encoding (BPE) technique [23], which involves splitting the words into more manageable subwords. Experiments are conducted using both unfiltered datasets and after the removal of mistranslation, obtained from standardization and tokenization using the Indic NLP library [36].

**Algorithm 1 :** Removal of Mistranslation
**Input**: Train dataset $CTR$, Validation dataset $CV$, Source Language $LS$, $CV|T$ denotes the part of $CV$ written in language T, Target Language $LT$, Parameter: $y$-Number of Epochs

1: $TF \leftarrow CTR + CV$
2: Train model on $CTR$ from $LS$ to $LT$ for $y$ epochs with best checkpoints saved in $P_B$
3: $C_V^0{}_{|T} \leftarrow P_B.\text{translate}\ (C_V{}_{|S})$
4: B ← BLEU($CV|T$, $CV0|T$)
5: Threshold ← B/4
6: **for** sentence pair ($P_S$, $P_T$) in ($C_{TR|S}, C_{TR|T}$) **do**
7: $P_T^0 \leftarrow P_B.\text{translate}\ (P_S)$
8: M ← BLEU ($P_T$, $P_T^0$)
9: **if** M $<$ Threshold **then**
10: Discard($P_S$, $P_T$) from $C_{TR}$
11:     **end if**
12: **end for**
13: Return $C_T$

**Table 1** Statistics of Dataset

| Dataset Division | Category | Pairs | Training Dataset |
|---|---|---|---|
| **Full dataset** | **Unfiltered** | ORI HIN | 990439 8431687 |
| | **After Removal of Mistranslation** | ORI HIN | 938582 5246667 |
| **Quarter dataset** | **Unfiltered** | ORI HIN | 247609 2115315 |
| | **After Removal of Mistranslation** | ORI HIN | 232255 1760096 |
| **500K Sentences dataset** | **Unfiltered** | ORI HIN | 500000 500000 |
| | **After Removal of Mistranslation** | ORI HIN | 473747 439207 |

**Table 2** Observed incorrect translations in the Dataset

| Language | Dataset Division | Wrong Translation Observed | Threshold |
|---|---|---|---|
| **ORI** | **Full** | 50860 | 1.729 |
| | **Quarter** | 14357 | 0.299 |
| | **500K** | 25256 | 1.009 |
| **HIN** | **Full** | 3184023 | 8.166 |
| | **Quarter** | 354222 | 7.505 |
| | **500K** | 59281 | 6.132 |

## 5 Results and Discussion

MT evaluation is the most important phase of any MT system. Three different evaluation metrics are utilized to verify the systems' effectiveness. These metrics are well-known and effective in determining the quality of translated texts. METEOR [38], RIBES [40], and BLEU [39] are the evaluation metrics used in this work. The evaluation uses the Flores-200 dataset [15] for testing. Table 3 displays the scores of the models trained with different sizes of dataset without any filtration as well as the removal of incorrect translation. RIBES and METEOR scores range from 0 to 1, whereas the BLEU score ranges from 0 to 100.

From the results, it is observed that Mistranslation Removal outperforms both Odia and Hindi languages. BLEU score is calculated only using the precision of the translation that is directly calculated over all n-grams (n ranges from 0 to 4) and recall is indirectly considered using the Brevity Penalty (BP) that penalizes shorter translations. METEOR score on the other hand makes use of precision and recall directly during calculation and penalizes a translation based on the number of chunks (fluent "chunks" of translations) it has, i.e., the lower the number of chunks, the lesser the penalty. In most cases, Mistranslation Removal gives a higher score. The BLEU scores achieved are higher than those of the unfiltered dataset. The removal of grave mistranslations improves dataset quality tremendously. The trade-off made between dataset quality and size is well paid off in the case of Odia wherein the loss in data is at most 6% while the increase in BLEU score is considerable. Results also reveal that Indic to the English language performs better than English to the English language.

### 5.1 ENG - ORI and ORI - ENG

In the case of the ORI dataset, models trained after the removal of mistranslation sentences from the dataset outperform the unfiltered dataset. The scores are found to be almost linearly increasing as the dataset size increases and this implies room for improvement to be proportional with the growth of ORI dataset. The lowest scores for the ORI dataset occur in the quarter-sized dataset. The lowest BLEU score is found to be 1.06 for the ORI-ENG language for the quarter dataset. METEOR and RIBES scores are also less for ENG-ORI language for the quarter dataset. For the ORI dataset, it becomes clear that using the full dataset division and removing mistranslation consistently produces positive results (i.e., 16.22 BLEU score) across a range of dataset division categories. From all the cases, the highest results in terms of metrics are achieved with the original, unfiltered dataset with a BLEU score of 7.06 and 16.22 for ENG-ORI and ORI-ENG, respectively. From all the cases, the 500k sentences unfiltered dataset for HIN has the lowest BLEU scores (24.54), while the full corpus after removing mistranslation has the highest BLEU scores (35.32). RIBES and METEOR scores after mistranslation removal from the full dataset for HIN-ENG are 0.53 and 0.84.

**Table 3 Evaluation Metrics with Unfiltered Dataset and After removal of mistranslation from the dataset**

| Dataset Division | Category | Pairs | BLEU | METEOR | RIBES |
|---|---|---|---|---|---|
| Full dataset | Unfiltered | ENG-ORI | 6.38 | 0.17 | 0.66 |
| | | ORI-ENG | 14.53 | 0.27 | 0.72 |
| | | ENG-HIN | 32.37 | 0.47 | 0.81 |
| | | HIN-ENG | 34.61 | 0.42 | 0.83 |
| | Mistranslation Removal | ENG-ORI | 7.06 | 0.21 | 0.68 |
| | | ORI-ENG | 16.22 | 0.33 | 0.73 |
| | | ENG-HIN | 32.45 | 0.50 | 0.81 |
| | | HIN-ENG | 35.32 | 0.53 | 0.84 |
| 500k Sentences dataset | Unfiltered | ENG-ORI | 3.48 | 0.11 | 0.61 |
| | | ORI-ENG | 9.15 | 0.20 | 0.65 |
| | | ENG-HIN | 24.58 | 0.39 | 0.77 |
| | | HIN-ENG | 26.75 | 0.40 | 0.79 |
| | Mistranslation Removal | ENG-ORI | 4.09 | 0.16 | 0.63 |
| | | ORI-ENG | 10.25 | 0.26 | 0.67 |
| | | ENG-HIN | 25.44 | 0.43 | 0.77 |
| | | HIN-ENG | 26.81 | 0.47 | 0.81 |
| Quarter dataset | Unfiltered | ENG-ORI | 1.06 | 0.06 | 0.46 |
| | | ORI-ENG | 3.95 | 0.13 | 0.56 |
| | | ENG-HIN | 30.03 | 0.44 | 0.79 |
| | | HIN-ENG | 32.94 | 0.45 | 0.83 |
| | Mistranslation Removal | ENG-ORI | 1.19 | 0.09 | 0.46 |
| | | ORI-ENG | 4.67 | 0.17 | 0.58 |
| | | ENG-HIN | 30.76 | 0.48 | 0.80 |
| | | HIN-ENG | 32.96 | 0.51 | 0.83 |

6. **Conclusion and Future Work**

This paper examines the impact of the dataset in terms of size and quality for the MT task. It also shows incorrect translations which are noticed in the dataset of two languages i.e. Hindi (HIN) and Odia. A baseline NMT model is built for both languages. The method to remove the wrong translation from the dataset is also presented in this paper. For all our experiments, various assessment metrics such as BLEU, RIBES, and METEOR are used to check the overall quality of translation. From the results, it is observed that Mistranslation Removal gives a higher score (in terms of evaluation metrics) than the original unfiltered dataset. It is also noticed that, even though the ILs-English and English-ILs systems are trained using the same corpus, ILs-English works more efficiently across all the evaluation metrics. Additionally, based on the analysis of our experiments, it is concluded that the size of the dataset is directly proportional to the quality of the translation. More language pairs with different dataset sizes for MT tasks need to be tested for future work. The impact of the removal of mistranslation with different variations of the threshold will be studied and investigated on other datasets to establish the underlying reasons behind the observed results.